\documentclass[10pt, conference, compsocconf]{IEEEtran}

\usepackage{cite}
\usepackage{amsmath,amssymb,amsfonts}
\usepackage{algorithmic}
\usepackage{graphicx}
\usepackage{multirow}
\usepackage{textcomp}
\usepackage{pgfplots}
\usepackage{makecell}
\usepackage{algorithm}
\usepackage{algorithmic}
\usepackage{dblfloatfix} 
\usepackage[caption=false]{subfig}
\usepackage{float}
\DeclareMathOperator*{\argmax}{argmax}
\def\BibTeX{{\rm B\kern-.05em{\sc i\kern-.025em b}\kern-.08em
    T\kern-.1667em\lower.7ex\hbox{E}\kern-.125emX}}
\newcommand{\widesim}[2][1.5]{
  \mathrel{\overset{#2}{\scalebox{#1}[1]{$\sim$}}}
}

\let\<\textless
\let\>\textgreater
\pgfplotsset{
    compat=1.8,
    tick align=inside,
    major tick length=2pt,
    legend style={font=\tiny, at={(0.5,1.02)}, anchor=south,legend columns=-1},
    tick label style={font=\scriptsize},
    axis line style=very thin,
    every axis/.append style={
        {font=\scriptsize},
    },
    width=\columnwidth,
    height=5.5cm,
}
\usepackage{hyperref}
\usepackage{cleveref}
\crefname{figure}{Fig.}{Figs.}
\crefname{table}{Table}{Tables}
\crefname{algorithm}{Algorithm}{Algorithms}
\crefname{equation}{Equation}{Equations}
\crefname{section}{Section}{Sections}

\begin{document}

\title{Generative Pre-training for Paraphrase Generation \\ by Representing and Predicting Spans in Exemplars}

\author{\IEEEauthorblockN{Tien-Cuong Bui\IEEEauthorrefmark{1},
Van-Duc Le\IEEEauthorrefmark{1},
Hai-Thien To\IEEEauthorrefmark{1} and
Sang Kyun Cha \IEEEauthorrefmark{2}}
\IEEEauthorblockA{\IEEEauthorrefmark{1}Dept. Electrical and Computer Engineering, Seoul National University, Seoul, South Korea\\ 
Email: cuongbt91@snu.ac.kr}
\IEEEauthorblockA{\IEEEauthorrefmark{2}Graduate School of Data Science, Seoul National University, Seoul, South Korea}
}

\maketitle

\begin{abstract}
Paraphrase generation is a long-standing problem and serves an essential role in many natural language processing problems. Despite some encouraging results, recent methods either confront the problem of favoring generic utterance or need to retrain the model from scratch for each new dataset. This paper presents a novel approach to paraphrasing sentences, extended from the GPT-2 model. We develop a template masking technique, named \textit{first-order masking}, to masked out irrelevant words in exemplars utilizing POS taggers. So that, the paraphrasing task is changed to predicting spans in masked templates. Our proposed approach outperforms competitive baselines, especially in the semantic preservation aspect. To prevent the model from being biased towards a given template, we introduce a technique, referred to as \textit{second-order masking}, which utilizes Bernoulli distribution to control the visibility of the first-order-masked template's tokens. Moreover, this technique allows the model to provide various paraphrased sentences in testing by adjusting the second-order-masking level. For scale-up objectives, we compare the performance of two alternatives template-selection methods, which shows that they were equivalent in preserving semantic information.

\end{abstract}


\section{Introduction}

Paraphrases are texts that express the same meaning but in different words. For instance, given a source sentence, \textit{``I want to be a businessman,"} we modify words or structures to provide a target sentence, \textit{``I'd like to be an entrepreneur."} Paraphrase generation is a long-standing challenge \cite{mckeown1983paraphrasing} in Natural Language Processing (NLP), which plays a crucial role in many problems such as question answering, text summarization, and information retrieval. Paraphrasing tools are essential to academic writing since numerous writers need such tools to avoid copyright problems. 

Traditional methods are mostly rule-based or statistical approaches \cite{bolshakov2004synonymous}, which substitute words by finding synonyms from lexical databases such as \cite{miller1995wordnet}. With the advances of the sequence-to-sequence (Seq2Seq) framework in NLP applications, many models have adopted this scheme in the paraphrase generation problem \cite{li2017paraphrase, gupta2018deep}. Despite some encouraging results, Seq2Seq-based models often encounter the problem of favoring generic utterance, especially when dealing with scarce data. In \cite{iyyer2018adversarial, chen2019controllable, kumar2020syntax}, the authors proposed controlling the paraphrasing process with encoded syntactic information using template sentences. This approach noticeably improves accuracy, especially in sentence structure modification. However, the methods mentioned above tend to precisely follow the structure of a given template sentence, which sometimes undermines creativity. 

Lately, pre-trained language models \cite{devlin2018bert, liu2019roberta, radford2018improving, radford2019language}, based on the Transformers \cite{vaswani2017attention}, achieve remarkable results in multiple NLP tasks. They construct language models in an unsupervised manner using either token masking methods or a generative approach. After pre-training with unlabeled text, the model can capture complex dependencies and understand natural language in various linguistic aspects from the token level to the semantic level. Among those, GPT-2 \cite{radford2019language} achieves excellent results in various text generation tasks. 

\begin{table}[t]
    \centering
    \caption{Examples of paraphrased sentences generated by SGCP \cite{kumar2020syntax}, CGEN \cite{chen2019controllable}, and ParafraGPT (Ours). Our generated outputs are semantically closer to inputs than other models.  \cite{chen2019controllable}}
    \begin{tabular}{c|ll}
        \hline
        \multirow{10}{*}{ParaNMT} & Source & -- they are the guys who hate fire! \\
        & Target & -- they hate fire! \\
        & CGEN & -- i like fire! \\
        & SGCP & -- i like fire. \\
        & (Ours) & -- they hate fire! \\
        \cline{2-3} 
        & Source & -- What he's gotta do is to dig a hole. \\
        & Target & -- He must dig a pit. \\
        & CGEN & -- I'll have a hole. \\
        & SGCP & -- I should have a hole. \\
        & (Ours) & -- He'll only dig that hole. \\
        \hline \hline
        \multirow{10}{*}{QQPPos}
        & Source & \makecell[l]{-- Do any animals other than humans commit \\ suicide because of emotional issues?}\\
        & Target & -- Do animals other than humans suicide?\\
        & CGEN & -- Do animals other than humans commit?\\
        & SGCP & -- Do animal animals commit in humans?\\
        & (Ours) & -- Do animals commit suicide in general? \\
        \cline{2-3}
        & Source & -- Why are there no ISIS attacks on Israel?\\
        & Target & -- Why isn't ISIS attacking Israel?\\
        & CGEN & -- Why are no Israeli issues Israel?\\
        & SGCP & -- Why aren't Pakistanis attack Israel?\\
        & (Ours) & -- Why doesn't Israel have terrorist attacks?\\
        \hline
    \end{tabular}
    \label{tbl:baseline_examples}
\end{table}

Inspired by these models' success, in this paper, we propose a method that utilizes the pre-trained GPT-2 model to tackle the paraphrase generation task, named ParafraGPT. The pre-trained GPT-2 model is fine-tuned with source-template-target triplets, including a source sentence, a template sentence, and a target sentence separated by special tokens. We found that guiding the model with templates during training can significantly enhance the accuracy of paraphrase generation. However, the model suffers redundant computations and negative effects from the attention of unnecessary words when utilizing full template sentences. Using the POS tagger of words, we substitute all nouns, adjectives, adverbs, and verbs (except modal verbs and ``to be" verbs in questions) of templates with a dummy token, as depicted in \cref{fig:masktempl}. We refer to this technique as \textit{first-order masking}. The paraphrase generation problem is changed to predicting arbitrary tokens to fill in the blanks in exemplars. Besides, we develop the \textit{second-order-masking} method as we want to prevent the model from adapting too much to a given template. Given a first-order-masked template, randomly selected tokens are ignored during training. This method is similar to the dropout regularization technique in deep learning models in the sense that the model can be trained with multiple templates for each source-target pair. A dropped-out token is selected using a Bernoulli distribution with probability of success $p$. We also call this probability the level of creativity of the model since it enables the model to generate various paraphrased sentences in testing with different $p$ values. For scaling up purposes, we implement two template selection methods: a tree-edit-distance-based (TED-based) method and an embedding-based algorithm. The TED-based approach is first proposed in \cite{kumar2020syntax}, whereas the idea of using sentence embeddings is encouraged by \cite{kazemnejad2020paraphrase}. Even though these two methods' performance is comparable, the embedding-based one is more practical in real-world applications due to its lightning-fast speed.

The main contributions of this paper are three-fold:
\begin{itemize}
    \item We propose a novel and efficient approach for paraphrase generation by training the GPT-2 model to complete masked templates. Our proposed method outperforms competitive baselines, especially in the semantic preservation metric. 
    \item We present the second-order-masking technique to prevent the model from being biased towards templates. This technique also allows the model to provide various paraphrased sentences corresponding to different masking probability values. 
    \item We implement two alternative methods to select templates for scale-up purposes. Besides, we elaborately analyze the influences of template selection and the second-order-masking level to output sentences in numerous experiments. 
\end{itemize}

The remainder of this paper is structured as follows. In \cref{sec:related_work}, we introduce the related work. The structure and technical details of our proposed method are presented in \cref{sec:proposed_method}. In \cref{sec:experimental_settings}, experimental settings are demonstrated. Then we presented experimental results in \cref{sec:experimental_results}. Finally, we conclude our work and discuss future research directions in \cref{sec:conclusion}.

\section{Related Work} \label{sec:related_work}
\subsection{Paraphrase Generation}
Paraphrase generation is a long-standing problem, which dates back to 1983 in \cite{mckeown1983paraphrasing}. Traditional methods solve this problem by using either machine translation approaches \cite{quirk2004monolingual, zhao2008combining, wubben2010paraphrase} or lexical databases \cite{bolshakov2004synonymous, kauchak2006paraphrasing}, such as Wordnet \cite{miller1995wordnet} to substitute words. Lately, with the rapid advances in Deep Learning and Neural Networks, many efficient methods have been proposed \cite{gupta2018deep, fu2019paraphrase}. These approaches aim to generate an output that is semantically similar to the input. However, most of them are built based on the encoder-decoder framework with LSTM networks, which must be retrained from scratch with new datasets.

\subsection{Controlled Paraphrasing Methods}
Many approaches based on syntactic and semantic information integration have been proposed to tackle the paraphrase generation problem \cite{iyyer2018adversarial,chen2019controllable}. The key idea of these methods is to control the output with a syntactic tree provided by a template sentence. Kumar et al. \cite{kumar2020syntax} have the same approach and propose a model that encodes both syntactic information using a tree encoder and semantic information using LSTM-based networks. Even though this model works well for clean datasets, it is inefficient with noisy datasets, wherein sentences contain grammatical mistakes and ill-formed structures. As opposed to them, our model exploits the capability of attention layers in GPT-2 to paraphrase the input sentence with respect to masked templates. Furthermore, our model does not require an additional tree encoder component.

Kazemnejad et al. \cite{kazemnejad2020paraphrase} adopt a retriever-editor architecture to paraphrase an input sentence. The retriever first selects the most similar source-target pair based on the embedding distance with the source. Then the editor modifies the input accordingly based on the Transformers \cite{vaswani2017attention}. Although this approach can be promising to handle both structure modification and word substitution in paraphrase generation, the model needs to be trained from scratch. Therefore, it is difficult to deal with low resources. In contrast, we hypothesize that pre-trained language models can be controlled to solve the paraphrasing task with both abundant data and low-resource situations.

\subsection{Fine-tuning Pre-trained Language Models}
GPT-2 \cite{radford2019language} has shown its capability to understand language through a generative pre-training period on the enormous open text. Witteveen et al. \cite{witteveen2019paraphrasing} and Hedge et al. \cite{hegde2020unsupervised} et al. propose a way to exploit large language models for paraphrase generation. These methods are similar to our \textit{na\"ive sample} configuration, which fine-tunes the GPT-2's weights using input samples with the form \textit{\<sos\> source\_tokens \<sep\> target\_tokens \<eos\>}. However, we observed many drawbacks of this approach, such as the source-replication problem, carefully described in \cref{evaluation_section}. 

\begin{figure*}[ht]
\centering
 \subfloat[Sample construction methods]{%
    \includegraphics[clip,trim=0.09cm 2cm 0.1cm 0.16cm,width=0.536\linewidth]{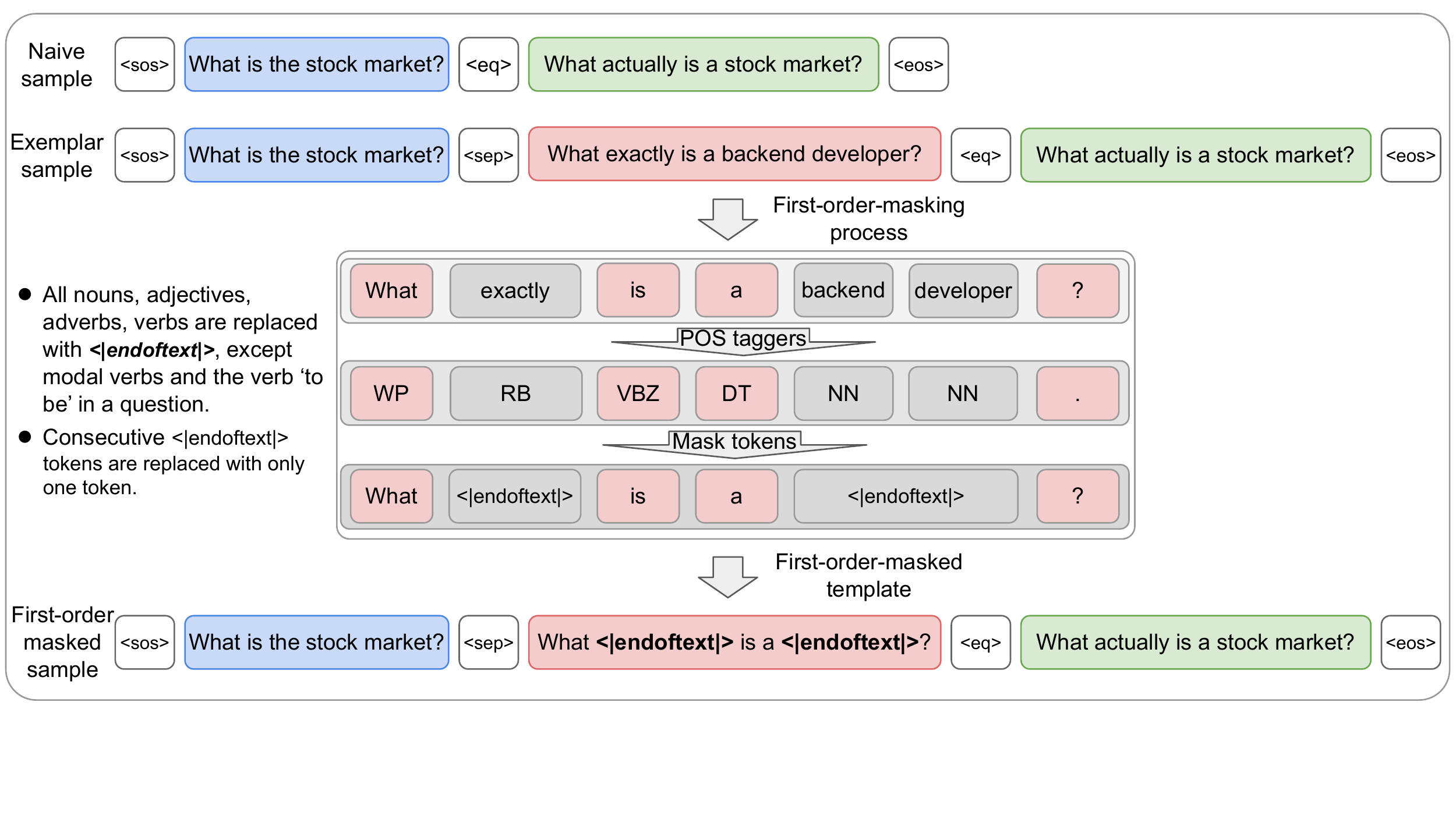}
    \label{fig:masktempl}
 }
 \hfill
 \subfloat[Training ParafraGPT with constructed samples]{%
    \includegraphics[clip,trim=0.15cm 1.15cm 0.15cm 1.0cm,height=4.56cm, width=0.43\linewidth]{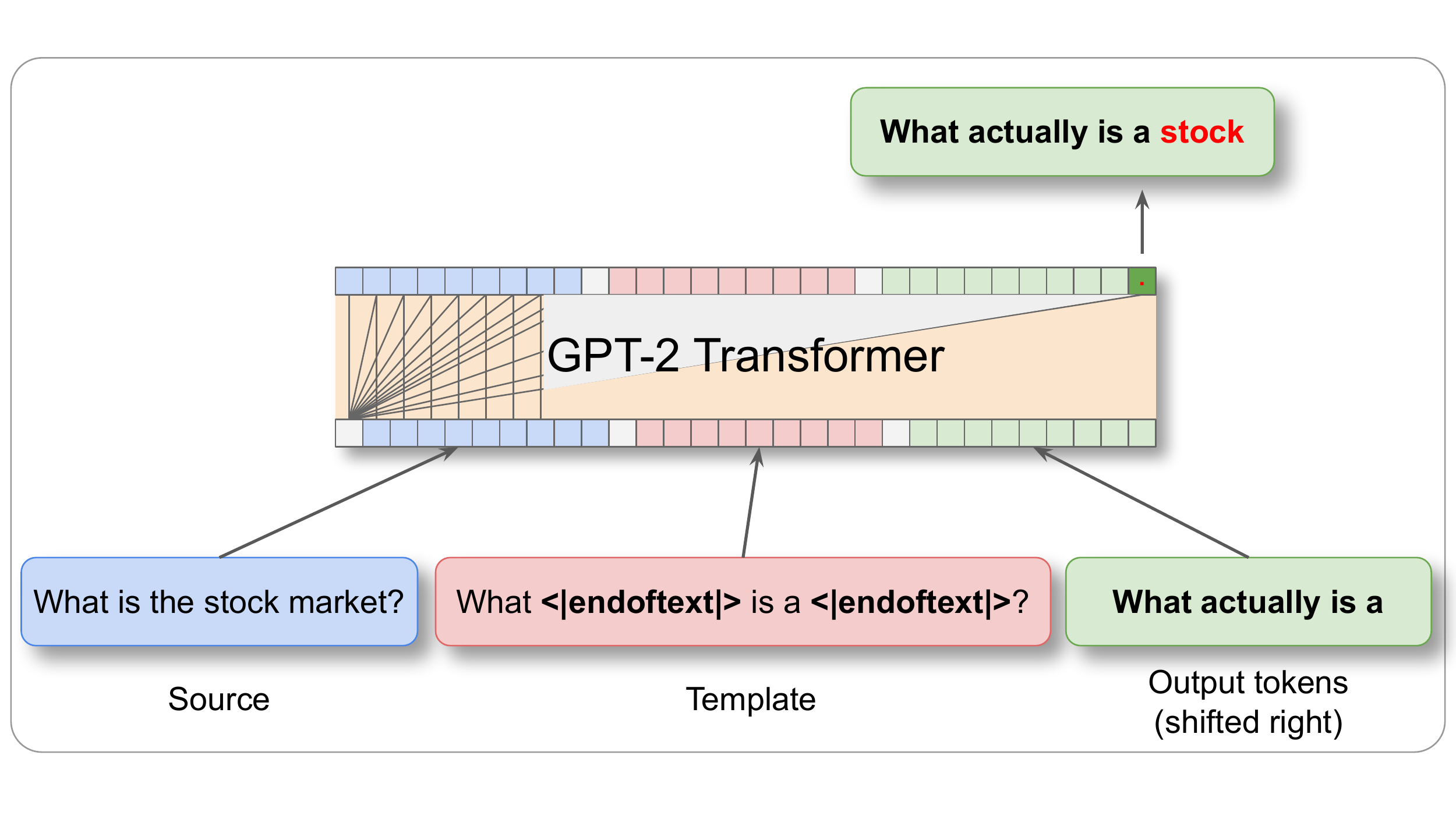}
    \label{fig:gpt2_finetune}
 }
 \hfill
\caption{Sample construction methods and Training ParafraGPT. We present three alternatives to construct input samples for training and testing. The model is based on pre-trained GPT-2 extended from Huggingface \cite{Wolf2019HuggingFacesTS}. It is trained with masked exemplar samples to learn how to paraphrase a source sentence given a masked template. The masked template is generated by replacing all nouns, adjectives, adverbs, and verbs (except modal verbs and the ``to be" verbs in a question), with a special token of the GPT-2 tokenizer \<$|$endoftext$|$\>.}
\label{fig:main_figure}
\end{figure*} 

\section{Proposed Method} \label{sec:proposed_method}
\subsection{Problem Formulation}

Given a source sentence $x$ and a template sentence $e$, a paraphrase generation model substitutes words or changes the order of words accordingly to output a target sentence $y$. The target sentence must have the same meaning as the source one but is expressed differently. Let $\mathcal{D} = \{x_i, e_i, y_i\}_{i=1}^N$ denotes a dataset which consists of a set of source-template-target triplets. Our goal is to train a model that maximizes $\Pi_{i=1}^N p_{\textrm{model}}(y_i|x_i, e_i)$. 
\subsection{Model Architecture}

We trained our ParafraGPT model based on the GPT-2 \cite{radford2019language} and the Huggingface Pytorch transformer \cite{Wolf2019HuggingFacesTS}. The GPT-2 adopts the decoder module of the original transformer architecture \cite{vaswani2017attention} and is trained on an enormous amount of web-text data. It treats the language modeling task in the generative style. Given a sequence of input tokens, the model tries to predict the next one with a stack of masked multi-head self-attention layers. OpenAI GPT-2 uses byte pair encodings as the tokenizer, wherein it maintains the token case sensitivity. A transformer-based model can efficiently model syntactic and grammatical relationships between words due to the capability of self-attention layers.

We construct the training samples for our model, as depicted in \cref{fig:main_figure}. Since a source and a target sentence are almost identical, applying the pre-trained GPT-2 to paraphrase generation is not straightforward. When we push to the model input sequences similar to the \textit{na\"ive sample} in \cref{fig:masktempl}, it quickly remembers templates and replicates them when making predictions. Using templates to control the paraphrase generation process is a common approach, as presented in many papers such as \cite{chen2019controllable, kumar2020syntax, kazemnejad2020paraphrase}. Inspired by that, we append a template sentence, which structure is close to the target, to the source. We denote this type of sample as the \textit{Exemplar Sample} in \cref{fig:masktempl}. 

\subsection{Template Masking and the Level of Creativity} \label{template_span_masking}

\noindent\textbf{First-order masking.} The \textit{Exemplar Sample} version in \cref{fig:masktempl} has two drawbacks. First, it causes a longer computation time during training. Second, unnecessary words in template sentences may have negative impacts on the adjustment process of attention layers. Therefore, using POS taggers, we substitute all nouns, adjectives, adverbs, and verbs (except modal verbs and ``to be" verbs in questions) of a template with \<$|$endoftext$|$\>. This process is presented at the bottom part of \cref{fig:masktempl}. By doing that, we transform the paraphrase generation task into a problem of predicting spans in exemplars using arbitrary tokens.

\noindent\textbf{Second-order masking.} We found that the model can be biased towards exemplars in the generation process. Therefore, it sometimes copies words from a template directly to a target sentence, causing grammatical mistakes or weird expressions. To prevent this issue, we control the visibility of tokens in a template with a probability $p$ obeyed the Bernoulli distribution. Each token in a first-order-masked template with the length $l$ is randomly substituted with \<$|$endoftext$|$\> by an indicator $I \in \{0,1\} \widesim{i.i.d} Bernoulli(p)$. $I = 1$ indicates that the model should replace the token, and $I = 0$ otherwise. Let $X$ be a random variable representing the number of dropped-out tokens in $l$ tokens following the binomial distribution with the probability $p \in [0,1]$. The probability of $k$ tokens being hidden out in $l$ independent Bernoulli trials is as follows:
\begin{equation}
    P(k,l,p) = P(X = k) = \binom{l}{k}p^k(1-p)^{l-k},
\end{equation}
for $k = 0,1,2,...,l$. The probability $p$ is referred to as \textit{the level of creativity} since it enables the model to output various paraphrased sentences with different values. It is usually small in training and can be set before the model execution. 

\subsection{Template Selection} \label{method_template_selection}
A paraphrasing system has to be able to scale up to serve the requests of numerous users. Therefore, the template selection process needs to be lightning fast. Our work implements two alternatives for template selection a tree-edit-distance-based method and an embedding-base one.

\noindent\textbf{TED-based method.} In \cite{kumar2020syntax}, the authors propose using the tree-edit distance to select template sentences. However, only templates for the validation and test set are pre-computed since performing this process on a large number of training samples takes a long time. For the training set, target sentences are directly used as templates. This method consists of three steps. First, a set of all sentences in the test and dev set is constructed, which is approximately 12K sentences. For each source-target pair in these two sets, a set of template candidates is selected as the following rules: 
\begin{enumerate}
    \item The sentence is neither the source nor the target sentence of the pair.
    \item The length difference (= number of tokens) between the candidate and the target sentence is at most two.
    \item The BLEU score between the target and the candidate must be smaller than 0.6.
\end{enumerate}
Finally, the candidate with the smallest TED is selected as the template. 

\begin{algorithm}
  \caption{Embedding-based Template Selection}
  \hspace*{\algorithmicindent} \textbf{Input}: source-target pairs  \\
  \hspace*{\algorithmicindent} \textbf{Output}: templates \\
  \begin{algorithmic}[1]
  \STATE{$\mathcal{D} = \{\}$} \COMMENT{a unique embedding set}
    \FOR{each sentence $s$ in dataset}
     \IF{$s$ not in $\mathcal{D}$}
      \STATE{$v_s$ = getSentenceEmbedding($s$) }
      \STATE{Add $v_s$ to $\mathcal{D}$}
     \ENDIF
    \ENDFOR
    \STATE {Build embedding index $\mathcal{I}\ \textrm{for}\ \mathcal{D}$}
    \STATE{$\mathcal{E} = \{\}$} \COMMENT{a returned template set}
    \FOR{all source-target pair \{$x,y$\} in a query set}
     \STATE{$v_x$ = getSentenceEmbeddingFromD($x$) }
     \STATE{$k\_sample$ = searchNeighbors($\mathcal{I}$, $v_x$)}
     \STATE{Remove $x$ and $y$ in $k\_sample$}
     \STATE{Randomly select a sentence $\hat{k}$ from $k\_sample$}
     \STATE{Add $\hat{k}$ to $\mathcal{E}$}
    \ENDFOR
    \STATE{\textbf{return} $\mathcal{E}$}
  \end{algorithmic}
\label{templt_embalg}
\end{algorithm}

\noindent\textbf{Embedding-based method.} Practically, the TED-based method is slow due to the computational overhead of the tree-edit distance on a large set. Kazemnejad et al. \cite{kazemnejad2020paraphrase} suggest using sentence embeddings and the cosine distance to find relevant samples for an input sentence. Inspired by that, we design an algorithm to select templates for the train, dev, and test set, as presented in \cref{templt_embalg}. The algorithm is practical in real-world applications with the advances in Sentence Embedding \cite{reimers2019sentence} and the indexing technique \cite{JDH17}.

\subsection{Training}
We construct the problem as a regular language modeling problem. Therefore, we aim to maximize the log-likelihood objective function

\begin{equation}
    \mathcal{L} = \sum_{(x,e,y)\in \mathcal{D}}\sum_{t=1}^T \textrm{log}\:P(y_t|y_{t-1}, x,e;\theta),
\end{equation}
where $t$ is the current step, $T$ is the number of generated tokens, and $\theta$ denotes ParafraGPT model parameters. Then, we formulate an optimization problem as follows:
\begin{equation}
    \theta^* = \argmax_\theta \mathcal{L}(\theta).
\end{equation}

\section{Experimental Settings} \label{sec:experimental_settings}

\begin{table*}[htbp] 
    \centering
    \caption{Results on the QQP-Pos and ParaNMT-small dataset. Source-as-Output and Exemplar-as-Output are used only for indicator purposes. Numbers in bold and underlined typeface mean the best and second-best results, respectively.}
    \begin{center}
        \resizebox{\textwidth}{!}{
        \begin{tabular}{l|ccccc|ccccc}
            \hline
                \multirow{2}{*}{Models}&\multicolumn{5}{c|}{ParaNMT-small}&\multicolumn{5}{c}{QQP-Pos} \\\cline{2-11} 
            
              & BLEU & ROUGE-1 & ROUGE-2 & ROUGE-L & BERT SCORE & BLEU & ROUGE-1 & ROUGE-2 & ROUGE-L & BERT SCORE \\
             \hline \hline
             Source-as-Output & 18.5 & 50.6 & 23.1 & 47.7 & 86.2 & 17.2 & 51.9 & 26.2 & 52.9 & 84.9 \\
            Exemplar-as-Output & 3.3 & 24.4 & 7.5 & 29.1 & 74.2 & 16.8 & 38.2 & 20.5 & 43.2 & 78.2 \\
            \hline
            SCPN \cite{iyyer2018adversarial} (2018) & 6.4 & 30.3 & 11.2 & 34.6 & 73.7 & 15.6 & 40.6 & 20.5 & 44.6 & 77.6 \\
            CGEN \cite{chen2019controllable} (2019) & 13.6 & 44.8 & 21.0 & 48.3 & 79.5 & 34.9 & 62.6 & 42.7 & 65.4 & 85.8 \\
            SGCP \cite{kumar2020syntax} (2020) & \textbf{15.3} & 46.6 & \underline{21.8} & \underline{49.7} & 80.5 & \textbf{36.7} & \textbf{66.9} & \textbf{45.0} & \textbf{69.6} & 87.2 \\
            \hline
             Na\"iveGPT & 7.3 & \underline{47.0} & 16.7 & 46.6 & \underline{80.7} & 15.8 & 52.1 & 25.3 & 51.7 & 84.5 \\
             \hline
             ParafraGPT-U & \underline{14.5} & \textbf{49.6} & \textbf{21.9} & \textbf{50.9} & \textbf{82.4} & 34.7 & 64.1 & 40.6 & 66.4 & 88.2 \\
             ParafraGPT-C & - & - & - & - & - & 35.6 & 66.1 & 43.5 & 68.3 & \underline{88.5} \\
             ParafraGPT-UC & - & - & - & - & - & \underline{35.9} & \underline{66.7} & \underline{43.7} & \underline{68.9} & \textbf{88.9} \\

            \hline 
        \end{tabular}
        }
    \end{center}
    \label{tbl:comparison_baselines}
\end{table*}

\subsection{Datasets}
We refer to \cite{kumar2020syntax} and select two datasets for experiments.

\noindent\textbf{QQP-Pos.} It is extracted from the Quora Question Pairs (QQP) dataset, which includes 400K source-target pairs. Each pair is labeled as negative if two questions are not duplicates and positive otherwise. Specifically, it has 150K positive and 250K negative pairs. Only 150K positive pairs are selected for experimental purposes. In \cite{kumar2020syntax}, sentences are lowercased and removed special tokens. Unlike them, we conduct experiments on both the original (case sensitive) and the pre-processed data. We denote these two cases as ParafraGPT-C and ParafraGPT-U, correspondingly. We also conduct an experiment that first trains the model on the pre-processed data then on the original data and refers to this setting as ParafraGPT-UC. As similar to \cite{kumar2020syntax}, 140K source-target samples are selected as the training set, 3K pairs for each test and dev set.

\noindent\textbf{ParaNMT-small \cite{chen2019controllable}.} It consists of approximately 500K source-target pairs. Specifically, it includes 1300 source-template-target manually annotated samples, divided into 800 for test data and 500 for the validation set. As similar to \cite{kumar2020syntax, chen2019controllable}, we use the target sentence of a source-target pair in training set as the template during training. Furthermore, we do not conduct experiments with the embedding-based template selection method on this dataset.

ParaNMT-small is a part of the ParaNMT-50M dataset \cite{sennrich2015neural}, generated through the back-translation of English sentences. The dataset has noise since many sentences are not grammatically correct. Therefore, we hypothesize that GPT-2 with the pre-trained phase on massive open-text data can solve the grammatical problem of the generated sentences.

\subsection{Baselines}
As similar to \cite{kumar2020syntax}, we report the evaluation scores of direct-copied outputs from either sources or templates. 
\begin{itemize}
    \item \textbf{Source-as-Output}: The output is copied directly from the input.
    \item \textbf{Exemplar-as-Output}: Template sentences are outputs.
\end{itemize}
Next, we compare our proposed approach with other methods that share the same template adoption idea.
\begin{itemize}
    \item \textbf{SCPN} \cite{iyyer2018adversarial} is a Seq2Seq-based model, including two LSTM encoders to process syntax and semantics. The decoder is also LSTM-based networks combined with a copying mechanism \cite{see2017get} and soft-attention \cite{bahdanau2014neural} over the encoder states. 
    \item \textbf{CGEN} \cite{chen2019controllable} also constructs two encoders to handle semantic and syntactic inputs. The model is trained with a multi-task paradigm, wherein the objective loss function combines both a word position loss and a paraphrase generation loss. We compare our model's results with their best setting.
    \item \textbf{SGCP} \cite{kumar2020syntax} combines syntax information provided by template trees with semantic information of the source. It has two configurations, wherein one uses full constituency parse tree information, whereas the other manipulates information of different tree levels. The latter can be considered as an ensemble method. We compare our model's results with the former setting for a fair comparison since our results are generated from a greedy technique.
    \item \textbf{Na\"iveGPT}: We fine-tune the pre-trained GPT-2 model \cite{radford2019language, Wolf2019HuggingFacesTS} with the \textit{Na\"ive Sample} version in \cref{fig:masktempl}.

\end{itemize}
To compare with these baselines, we report our model's results of three configurations as follows:
\begin{itemize}
    \item \textbf{ParafraGPT-U}: it is a setting that we train the model with the pre-processed data of SGCP, which are removed special tokens and lowercased.
    \item \textbf{ParafraGPT-C}: the model is trained with the original QQP-Pos dataset, wherein we select the test and dev set exactly the same as SGCP.
    \item \textbf{ParafraGPT-UC}: we continue training the model on the case sensitive data similar to the ParafraGPT-C after training it 50 epochs with the ParafraGPT-U setting.
\end{itemize}

\subsection{Evaluation} \label{evaluation_section}
In the paraphrase generation task, human evaluation is critical to measure the performance of a generation model. However, it usually requires both time and budgets to complete the evaluation. Therefore, we use the following two types of metrics to evaluate the results of all methods.
\begin{itemize}
    \item \textbf{Alignment-based metrics:} We first evaluate all models on four metrics BLEU, ROUGE-1, ROUGE-2, and ROUGE-L. 
    \item \textbf{Semantic-based metric:} The BERT score \cite{zhang2019bertscore} exploits meaningful contextualized token embeddings of pre-trained BERT-based models. It calculates the similarity of two sentences by summing up cosine similarities between the embeddings of their tokens. \textit{N}-gram-based metrics fail to match paraphrases since both words and the source sentence structure can be changed. We use the pre-trained model \textit{roberta-large-mnli} \cite{liu2019roberta} as the embedding function.
   
\end{itemize}
\subsection{Setup}
\noindent\textbf{Pre-processing.} For the TED-based approach, we referred to \cite{kumar2020syntax} and reused its pre-computed templates. To implement \cref{templt_embalg}, we utilized Sentence Transformers \cite{reimers2019sentence} as the sentence embedding method and Faiss \cite{JDH17} as the embedding indexing technique. Both libraries work extremely fast and can easily support a large-scale system. We used Stanza \cite{qi2020stanza} to find POS taggers of tokens. Based on the POS information, we constructed first-order-masked samples. These processes were done before training to prevent computational costs of redundant operations. 

\noindent\textbf{Implementation details.} The Huggingface \cite{Wolf2019HuggingFacesTS} library provides a complete solution to load pre-trained weights for fine-tuning purposes. Due to the GPU memory limitation, our model extends the smallest version of GPT-2, which has 12 heads and 117 million parameters. We fail to load a bigger version of GPT-2, such as the medium one and the large one into our GPUs. The number of parameters of them is 3x and 7x more than the base model, respectively. 

We kept all hyper-parameters, such as the dropout rate as the original configurations. The learning rate was set as 6.25e-5. We used the Adam optimizer with $\beta1=0.9$ and $\beta2=0.999$. We did not use warm-up steps and directly trained the model until it cannot increase validation accuracy. On the QQP-Pos dataset, the model converged after approximately 150 training epochs. We then loaded the model checkpoint from ParafraGPT-U to train on the ParaNMT-small dataset around 20 epochs instead of training from scratch. During training, we set the teacher forcing ratio as $0.9$ and the second-order-masking probability $p$ as $0.15$. In testing, all results were generated using a greedy method for fast computation, and $p$ was set as $0.0$ in all experiments except in \ref{sec:template_selection}. 

We utilized the TED-based approach in all experiments, except in \cref{sec:template_selection}. All analyzes were conducted with the QQP-POS dataset.


\section{Experimental Results} \label{sec:experimental_results}

\subsection{Comparing with Baselines}
As reported in \cref{tbl:comparison_baselines}, we compare our model's results with competitive baselines on two datasets QQP-Pos and ParaNMT-small. We aim at assessing the capability of all methods in syntactic modification and semantic preservation. In alignment metrics, our model outperforms SCPN (Iyyer et al., 2018 \cite{iyyer2018adversarial}) and CGEN(Chen et al., 2019 \cite{chen2019controllable}) in all scores. Its scores are comparable to SGCP (Kumar et al., 2020) on the QQP-Pos dataset, whereas it performs better than SGCP on the ParaNMT-small dataset, which is much more challenging. Specifically, ParafraGPT is superior to all competitors in the BERT score, highlighting the capability of our method in semantic preservation. 

We observed that Na\"iveGPT tends to replicate a source sentence in the generated output, resulting in the similarity of its evaluation scores with the Source-as-Output, especially on QQP-Pos. Without template guidance, the GPT-2 model cannot differentiate a source and a target sentence due to the high similarity between them.  Therefore, it cannot specify which words are needed to substitute in testing or change the word order accordingly. Although the replication problem was relieved on the ParaNMT-small dataset, the generation accuracy did not increase considerably.

\subsection{The Level of Creativity and Template Selection} \label{sec:template_selection}

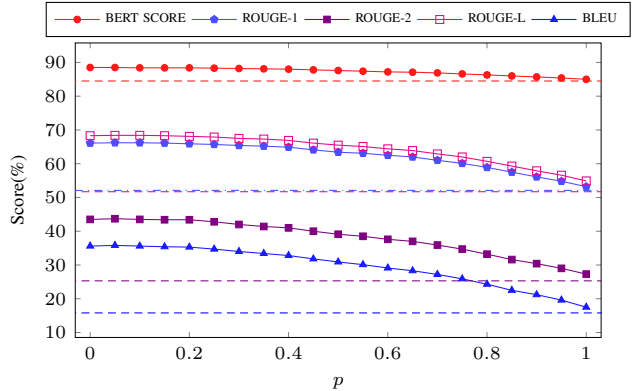
\begin{figure}[ht]
    \centering
    \begin{tikzpicture}
    \begin{axis}[
        mark size=1.25pt,
        xmin=-0.03,xmax=1.03, 
        ytick={0,10,20,30,40,50,60,70,80,90,100},
        legend style={},
        ylabel=Score(\%),
        xlabel=$p$,
        ]
        \addplot[color=red!90!white,mark=otimes*] coordinates{
            (0.0,88.5)
            (0.05,88.5)
            (0.1,88.4)
            (0.15,88.4)
            (0.20,88.4)
            (0.25,88.3)
            (0.30,88.2)
            (0.35,88.1)
            (0.40,88.0)
            (0.45,87.8)
            (0.50,87.6)
            (0.55,87.4)
            (0.60,87.2)
            (0.65,87.1)
            (0.70,86.9)
            (0.75,86.6)
            (0.80,86.3)
            (0.85,86.0)
            (0.90,85.7)
            (0.95,85.4)
            (1.0,85.0)
        };
        \addlegendentry{BERT SCORE}
    
        \addplot[color=blue!70!white,mark=pentagon*, mark options={mark size=1.5pt}] coordinates {
            (0.0,66.1)
            (0.05,66.2)
            (0.1,66.2)
            (0.15,66.1)
            (0.20,65.9)
            (0.25,65.7)
            (0.30,65.4)
            (0.35,65.2)
            (0.40,64.9)
            (0.45,64.1)
            (0.50,63.4)
            (0.55,63.1)
            (0.60,62.5)
            (0.65,62.0)
            (0.70,61.0)
            (0.75,60.1)
            (0.80,58.9)
            (0.85,57.5)
            (0.90,56.1)
            (0.95,54.8)
            (1.0,53.2)
        };
        \addlegendentry{ROUGE-1}
        \addplot[color=violet,mark=square*] coordinates{
            (0.0,43.5)
            (0.05,43.7)
            (0.1,43.5)
            (0.15,43.4)
            (0.20,43.4)
            (0.25,42.8)
            (0.30,42.0)
            (0.35,41.4)
            (0.40,41.0)
            (0.45,40.0)
            (0.50,39.1)
            (0.55,38.5)
            (0.60,37.6)
            (0.65,37.0)
            (0.70,35.9)
            (0.75,34.7)
            (0.80,33.2)
            (0.85,31.6)
            (0.90,30.4)
            (0.95,29.0)
            (1.0,27.3)
        };
        \addlegendentry{ROUGE-2}
        \addplot[color=magenta,mark=square, mark options={mark size=1.5pt}] coordinates{
            (0.0,68.3)
            (0.05,68.4)
            (0.1,68.4)
            (0.15,68.3)
            (0.20,68.1)
            (0.25,67.9)
            (0.30,67.5)
            (0.35,67.3)
            (0.40,66.9)
            (0.45,66.1)
            (0.50,65.5)
            (0.55,65.1)
            (0.60,64.4)
            (0.65,63.9)
            (0.70,62.9)
            (0.75,62.0)
            (0.80,60.7)
            (0.85,59.3)
            (0.90,57.9)
            (0.95,56.6)
            (1.0,54.9)
        };
        \addlegendentry{ROUGE-L}
        \addplot[color=blue!90!white,mark=triangle*, mark options={mark size=1.5pt}] coordinates{
            (0.0,35.6)
            (0.05,35.8)
            (0.1,35.6)
            (0.15,35.4)
            (0.20,35.3)
            (0.25,34.7)
            (0.30,34.0)
            (0.35,33.4)
            (0.40,32.8)
            (0.45,31.8)
            (0.50,30.9)
            (0.55,30.1)
            (0.60,29.1)
            (0.65,28.3)
            (0.70,27.2)
            (0.75,25.9)
            (0.80,24.3)
            (0.85,22.5)
            (0.90,21.2)
            (0.95,19.6)
            (1.0,17.5)
        };
        \addlegendentry{BLEU}
        \addplot[mark=none, red!90!white, densely dashed] {84.5};
        \addplot[mark=none, blue!70!white, dashdotted] {52.1};
        \addplot[mark=none, violet, densely dashed] {25.3};
        \addplot[mark=none, magenta, densely dashed] {51.7};
        \addplot[mark=none, blue, densely dashed] {15.8};
    \end{axis}
    \end{tikzpicture}
    \caption{Results of different values of second-order-masking probability. Each template token is randomly masked out with respect to a probability $p$ drawn from the Bernoulli distribution. Dashed lines are the Na\"iveGPT's scores.}
    \label{fig:masking_control}
\end{figure}

\noindent\textbf{The level of creativity.} We present the results of 21 $p$ values in \cref{fig:masking_control}. We gradually increase the value of $p$ by $0.05$ from 0 to 1. When $p = 1$, the model does not use any information of a template, whereas when $p = 0$, it exploits all hints from the exemplar sentence.

As shown in \cref{fig:masking_control}, scores of all alignment metrics noticeably decrease as $p$ increases, emphasizing the importance of exemplars, wherein they work as guidance for the model to learn how to fill in the blanks to generate target sentences. Although the model's accuracy when $p = 1$ approaches the Na\"iveGPT performance (dashed lines), we observed that our model does not suffer from the source-replication problem. However, sometimes, the model substitutes words and completely changes the word order, thus hurting the precision. As shown in \cref{tbl:tgt_masking_sentences}, the model tends to generate sentences based on its creativity as the amount of information from templates is reduced. The results from this table also suggest that the model can provide multiple candidates by only adjusting the probability $p$ and no need for additional template sentences. 

\begin{table}[ht!]
    \centering
    \caption{Examples of paraphrased sentences corresponding to $p$.}
    \begin{tabular}{l|l}
        \hline 
        Source & Will the Electoral College vote for Hillary? \\
        Target & \makecell[l]{What are the chances of Electoral college votes for Hillary?} \\
        Template & \makecell[l]{What are the \<$|$endoftext$|$\> of  \<$|$endoftext$|$\> in \\ \<$|$endoftext$|$\>?} \\
        \hline
        $p = 0$ & Can the Electoral College elect Hillary Clinton? \\
        $p = 0.5$ & \makecell[l]{What are the chances of Electoral college votes Hillary \\ Clinton?} \\
        $p = 1$ & Does the electoral college have to vote for Hillary Clinton? \\
        \hline\hline
        Source & What is the best way to lose weight and not gain it back? \\
        Target & How can I lose weight slowly and naturally? \\
        Template & How can \<$|$endoftext$|$\> and \<$|$endoftext$|$\>? \\
        \hline
        $p = 0$ & How do I lose weight? \\
        $p = 0.5$ & How can I lose weight safely? \\
        $p = 1$ & How can I lose weight slowly and naturally? \\
        \hline 
    \end{tabular}
    \label{tbl:tgt_masking_sentences}
\end{table}

\begin{figure}[ht]
    \centering
    \begin{tikzpicture}
    \begin{axis}[
        ybar,
        legend cell align=left,
        area legend,
        enlarge x limits={abs=2cm},
        ylabel={Score (\%)},
        nodes near coords,
        every node near coord/.append style={font=\tiny},
        symbolic x coords={TED-based,Embedding-based},
        xtick=data,
        ]
    \addplot[fill=blue!30] coordinates{(TED-based,88.9) (Embedding-based,88.2)};
    \addplot coordinates{(TED-based,68.3) (Embedding-based,65.1)};
    \addplot[fill=yellow!30] coordinates{(TED-based,66.1) (Embedding-based,64.1)};
    \addplot[fill=green!30] coordinates{(TED-based,43.5) (Embedding-based,40.4)};
    \addplot[fill=gray!40] coordinates{(TED-based,35.6) (Embedding-based,34.9)};
    
    \legend{BERT SCORE,ROUGE-L,ROUGE-1,ROUGE-2,BLEU}
    \end{axis}
    \end{tikzpicture}
    \caption{Results of template selection methods. We compare the performance of TED-based and embedding-based template selection.}
    \label{fig:template_compare}
\end{figure}
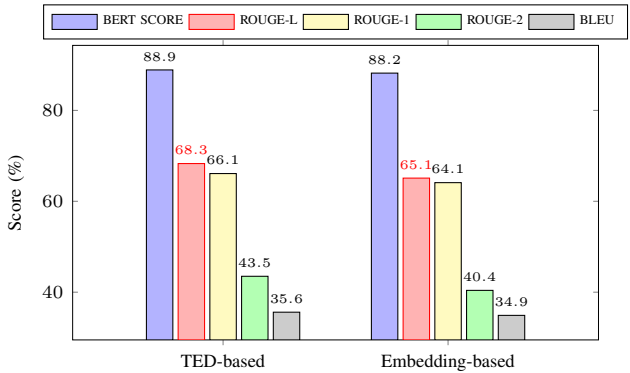

\noindent\textbf{Template selection methods.} As depicted in \cref{fig:template_compare}, the TED-based method's accuracy is superior to the embedding-based one, especially in alignment metrics. When using embedding-based templates, the model tends to generate sentences that are less close to the target in a syntactic structure due to a significant difference in syntax trees. Therefore, the alignment scores of the embedding-based approach are inferior to the TED-based one. However, both methods are comparable in the BERT score, which reveals that they are equivalent in the semantic preservation aspect. Specifically, we can allow the model to be more productive by combining the embedding-based template with second-order-masking levels in the generation process.

\subsection{Partial fine-tuning}
When fine-tuning all parameters, a language model provides the best performance in downstream tasks compared to partial fine-tuning some layers of the model, as reported in \cite{raffel2019exploring}. Two other alternative approaches for fine-tune an NLP model are adapter layers and gradual unfreezing. Although the former method is promising as it shows significant results in low-resource tasks, we could not make it work in this task. 

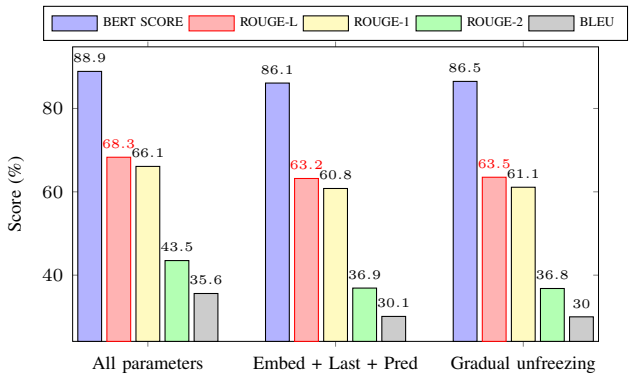
\begin{figure}[htbp]
    \centering
    \begin{tikzpicture}
    \begin{axis}[
        ybar,
        legend cell align=left,
        area legend,
        bar width=9pt,
        enlarge x limits={abs=1cm},
        ylabel={Score (\%)},
        nodes near coords,
        every node near coord/.append style={font=\tiny},
        symbolic x coords={All parameters,Embed + Last + Pred,Gradual unfreezing},
        xtick=data,
        ]
        \addplot[fill=blue!30] coordinates{(All parameters, 88.9) (Embed + Last + Pred, 86.1) (Gradual unfreezing, 86.5)};
        \addplot coordinates{(All parameters, 68.3) (Embed + Last + Pred, 63.2) (Gradual unfreezing, 63.5)};
        \addplot[fill=yellow!30] coordinates{(All parameters, 66.1) (Embed + Last + Pred, 60.8) (Gradual unfreezing, 61.1)};
        \addplot[fill=green!30] coordinates{(All parameters, 43.5) (Embed + Last + Pred, 36.9) (Gradual unfreezing, 36.8)};
        \addplot[fill=gray!40] coordinates{(All parameters, 35.6) (Embed + Last + Pred, 30.1) (Gradual unfreezing, 30.0)};
    
    \legend{BERT SCORE,ROUGE-L,ROUGE-1,ROUGE-2,BLEU}
    \end{axis}
    \end{tikzpicture}
    \caption{Results of partial fine-tuning the pre-trained model. For brevity, the embedding layer, the prediction layer, and the last attention layer are abbreviated to \textit{embed}, \textit{pred}, \textit{last}, respectively.}
    \label{fig:partial_training}
\end{figure}

The model has to be capable of selecting appropriate tokens to fill in the blanks in a template sentence to paraphrase a source sentence. Therefore, the prediction performance improves as proportional to the number of attention layers updated during training, as presented in \cref{fig:partial_training}. Unsurprisingly, ParafraGPT can achieve superior results compared to other configurations when it is retrained in all parameters. Besides, we tried to train only the embedding and the prediction layer, but the experiment did not work as expected and showed inferior results. Interestingly, when updating only the embedding layer, the last attention block, and the prediction layer, the model achieves comparable performance with the \textit{Gradual unfreezing} setting. It suggests that we can select an appropriate training strategy corresponding to specific requirements.

\subsection{Learning Cased or Uncased Tokens?}
As we present in \cref{tbl:comparison_baselines}, using case sensitive data for training improves the accuracy in both alignment metrics and the semantic one. GPT-2 is originally trained with cased tokens, thus causing its poor performance with the uncased dataset. It is due to the GPT-2 tokenizer does not recognize words like \textit{iphone} or \textit{itunes} as single words, so that it blocks the ParafraGPT-U model from using existing embeddings of these words. Therefore, dealing with case sensitive tokens allows GPT-2 to enhance the language model's semantic understanding, especially with names. 

Besides, we found that when retraining the ParafraGPT model on the case sensitive data after pre-training it with the lowercase version, the results were better than training with either one of them. The reason is that the QQP-Pos dataset contains numerous names or proper nouns that may not be exposed to the GPT-2 model during its pre-training phase. Therefore, the case-sensitive property significantly impacts the embedding updating process since the model has to learn to reconstruct names and weird words from the character-level. For instance, the word \textit{Urjit Patel} is tokenized as three tokens \textit{ĠUr}, \textit{jit}, and \textit{ĠPatel}, whereas it is separated into \textit{Ġur}, \textit{jit}, \textit{Ġp}, and \textit{atel} with lowercased tokens.

\section{Conclusion and Future Work} \label{sec:conclusion}
In this paper, we proposed a novel and practical approach to paraphrase sentences. First, the \textit{first-order-masking} method was implemented, which masks out selected words in templates using POS taggers. The model, extended from GPT-2, was trained to complete templates with appropriate tokens so that the output is close in meaning to the source sentence. Our proposed method outperformed competitive baselines, especially in the semantic score. Besides, we introduced an efficient technique, named \textit{second-order masking}, which eliminates dependence on templates. This technique also allows the model to provide various paraphrased sentences in testing, corresponding to different second-order-masking levels. We compared the accuracy of two alternatives for template selection: a TED-based approach and an embedding-based algorithm, and observed that they are comparable in the semantic preservation aspect. 

In future work, we aim to generalize the proposed approach to controllable text generation problems. Besides, ensemble methods are considerable since the model can generate various paraphrased sentences with different second-order-masking levels. Furthermore, since all of our results are based on a greedy technique, we can improve accuracy by exploring appropriate beam search settings.

\bibliographystyle{IEEEtran}
\bibliography{bibliography}

\end{document}